\documentclass[10pt]{article}

\usepackage[margin=1in]{geometry}
\usepackage{amsmath,amssymb,amsthm}
\usepackage{booktabs}
\usepackage{graphicx}
\usepackage{hyperref}
\usepackage{xcolor}
\usepackage{microtype}
\usepackage{multirow}
\usepackage{array}
\usepackage{enumitem}
\usepackage{algorithm}
\usepackage{algpseudocode}
\usepackage{caption}
\usepackage{subcaption}
\usepackage{capt-of}

\newcommand{\K}{\mathbf{K}}
\newcommand{\V}{\mathbf{V}}
\newcommand{\Q}{\mathbf{Q}}
\newcommand{\U}{\mathbf{U}}

\newcommand{\E}{\mathbb{E}}
\newcommand{\R}{\mathbb{R}}
\newcommand{\deff}{d_{\mathrm{eff}}}

\newcommand{\btail}{b_{\mathrm{tail}}}
\newcommand{\dtwo}{d_2}

\theoremstyle{plain}
\newtheorem{proposition}{Proposition}

\theoremstyle{definition}

\theoremstyle{remark}

\begin{document}

\title{\textbf{Ablation, Statistical Inference, and Validation for KV-Cache Compression}}

\author{
  Paolo D'Alberto\thanks{This work was developed in collaboration with Claude (Anthropic).} \and
  Ashish Sirasao \and
  Elliott Delaye \and
  Rajeev Patwari \\
  Advanced Micro Devices, Inc.\\
  \texttt{\{paolo.dalberto, ashish.sirasao, elliott.delaye, rajeev.patwari\}@amd.com}
}

\date{}
\maketitle

\begin{abstract}

We present a systematic comparative study of two families of KV-cache
compression schemes: TurboQuant (TQ), which applies a randomized
Walsh-Hadamard rotation and a data-oblivious Beta Lloyd-Max codebook, and
SpectralQuant (SQ), which calibrates a per-head eigenbasis and allocates
bits via water-filling.  Both families optionally append a 1-bit
Johnson-Lindenstrauss (QJL) residual sketch on the key path, the value path,
or both.

We make three contributions.  First, we perform a full ablation
across multiple QJL variants and embedding dimensions, and show that only
three schemes are non-dominated: scalar quantization without rotation,
WHT rotation with Beta Lloyd-Max codebook, and the latter augmented with
QJL on keys.  

Second, we introduce a statistical validation methodology for comparing
implementations: Python (oracle) and HIP/GPU (production) use different
random number generators and matrix operations as explicit experimental
variables.  We apply the Kolmogorov-Smirnov test to separate systematic
codec differences from implementation-induced variance, and identify the
K-path as a direct signature of Jensen's inequality amplifying score
variance through the softmax nonlinearity.

Third, we compare the final schemes across all regimes and dimensions
and derive regime-specific recommendations.  Heavy-tailed data is
catastrophic for any eigenbasis-based method: sample covariances are
destabilised by outliers, the calibrated basis is systematically
misaligned, and no budget increase recovers the loss.  On structured
regimes, SQ wins when separate K and V eigenbases provide genuine
compression; water-filling reduces to uniform allocation throughout.
We also characterize the self-calibrating nature of the effective semantic
dimension $\deff$, which adapts to the available calibration budget rather
than recovering the true data rank --- a property that explains both
surprising wins and non-monotone scaling behaviors.

\end{abstract}

\section{Introduction}
\label{sec:intro}

Transformer inference at scale is bounded by memory bandwidth: KV-cache access
dominates total memory traffic for long-context generation, and reducing cache
size directly translates to latency and throughput improvements.  Quantization
of keys and values is the standard approach, and a growing body of work shows
that aggressive quantization to 2--4 bits per element is possible without
meaningful accuracy loss~\cite{tq,kivi,kvquant}.

A recurring challenge in evaluating KV-cache compression is the difficulty of
attributing observed quality differences to specific algorithmic choices.
Evaluations on real large language model (LLM) traffic conflate distributional properties, hardware
effects, and algorithmic assumptions, making it hard to understand \emph{when}
and \emph{why} a method succeeds or fails.  This paper introduces a methodology
for controlled evaluation of KV-cache quantization schemes: a set of six
synthetic statistical regimes, each designed to isolate one structural
assumption of the compression pipeline, together with a statistical framework
for distinguishing systematic algorithmic differences from implementation noise.
The full evaluation is released as an open-source HIP/C++ benchmark targeting
AMD GPUs; any new compression scheme can be evaluated against the same regimes
by implementing a single scheme interface, without modifying the data generation,
metrics, or statistical validation infrastructure.

We instantiate the methodology on two representative families.
\textbf{TurboQuant (TQ)}~\cite{tq} is data-oblivious: a randomized
Walsh-Hadamard transform spreads each token's energy uniformly across all $d$
dimensions, making an analytical Beta-distribution codebook applicable
uniformly, and an optional 1-bit QJL residual sketch corrects the dominant
direction of the quantization error.
\textbf{SpectralQuant (SQ)}~\cite{sq_original} is data-adaptive: it calibrates
a per-head eigenbasis from representative tokens, concentrates bits on
high-variance semantic dimensions via water-filling, and corrects residuals
with a selective QJL sketch applied only to the top-$\deff$ dimensions.
These two families represent opposite ends of the data-dependence spectrum
and together exercise the full range of assumptions our regimes are designed
to test.

The synthetic regimes are not held-out test sets.  Their purpose is
diagnostic: to reveal the conditions under which a compression scheme achieves
its full potential or fails catastrophically, and to explain the mechanism
behind each outcome.  Each regime targets one assumption in isolation ---
distribution shape, K/V subspace alignment, or eigenvalue decay --- so that
failures can be attributed and understood rather than merely observed.

The main findings are: (1) a full ablation over eight QJL variants identifies
three non-dominated schemes, eliminating the rest; (2) the statistical
validation framework reveals that K-path QJL variance is exponentially
amplified by softmax (Jensen's inequality), while V-path variance is not ---
a distinction invisible to accuracy-only evaluations; (3) TQ dominates on
heavy-tailed data where eigenbasis calibration fails; (4) SQ wins on
structured regimes at sufficient budget, provided K and V are calibrated
on separate representative sets; and (5) water-filling reduces to uniform
allocation in all tested regimes, and the effective semantic dimension
$\deff$ self-calibrates to the available calibration budget rather than
recovering the true data rank.

\section{Background}
\label{sec:background}

Standard multi-head attention computes, for each head,
\begin{equation}
  T = \mathrm{softmax}\!\left(\frac{\Q\K^\top}{\sqrt{d}}\right)\V,
\end{equation}
where $\K,\V \in \R^{S \times d}$ are the key and value caches for $S$ past
tokens and $\Q \in \R^{N_q \times d}$ is the query batch.

\textbf{TurboQuant.}  Given per-head random signs $s \in \{-1,+1\}^d$, TQ
rotates each key as $y = \mathrm{WHT}(k \odot s) / \sqrt{d}$, making $y$
approximately Beta$((d{-}1)/2, (d{-}1)/2)$ distributed for any $k$ on the
unit sphere.  A single analytical Lloyd-Max codebook is fitted to this
distribution and shared across all dimensions.  Optionally, the
WHT-domain residual $e = y - \hat{y}$ is sketched using a randomized Hadamard transform
(RHT): $b = \mathrm{sign}(\mathrm{WHT}(e \odot s_2))$
for a fixed sign vector $s_2$, and the correction $\Delta \hat{y} =
\frac{\sqrt{\pi/2}}{d}\|e\|\,s_2 \odot \mathrm{WHT}(b)$ is added at decode.
\textbf{QJL on K} improves the attention score estimate
$\langle q, k\rangle$; \textbf{QJL on V} improves the value reconstruction
$\hat{V}$.  The inner product $\langle q, k\rangle$ is a sum over the embedding
dimension $d$, so the QJL sketch of $k \in \R^d$ is theoretically grounded
(Zandieh et al.\ 2024~\cite{qjl}).  The attention output $T_j = \sum_i a_i V_{ij}$
sums over tokens $S$, so a QJL sketch of $V_i \in \R^d$ does not target the
relevant inner product and is a heuristic correction.

\textbf{SpectralQuant.}
The core idea of SpectralQuant is a reallocation of bits from many
low-variance dimensions to a few high-variance ones.  Consider a head
with $d = 128$ dimensions and a budget of $b = 2$ bits: a uniform
scheme allocates $2 \times 128 = 256$ bits per token, spreading the
budget thinly across all dimensions regardless of their importance.
If the key vectors lie near a rank-4 subspace, only 4 directions carry
meaningful signal.  SQ identifies these directions from calibration data
and allocates, say, $8$ bits to each of the 4 semantic dimensions (32 bits
total), then quantizes the remaining 124 tail dimensions at 2 bits each
(248 bits), for a total of only $280$ bits --- comparable storage to the
uniform scheme, but with the critical dimensions quantized $4\times$ more
finely.  Equivalently, SQ can achieve the accuracy of a high-budget
uniform scheme at a fraction of the storage cost.

Given calibration keys $K_{\mathrm{cal}} \in
\R^{n_{\mathrm{cal}} \times d}$, SQ computes the empirical covariance
$C_K = K_{\mathrm{cal}}^\top K_{\mathrm{cal}} / n_{\mathrm{cal}}$, extracts
the top-$\deff$ eigenvectors $\U_K \in \R^{d \times \deff}$ via block power
iteration, and projects each key as $z_k = \U_K^\top k$.  A separate
eigenbasis $\U_V$ is calibrated from value tokens.  The effective dimension
is the participation ratio $\deff = \mathrm{round}(\mathrm{tr}(C)^2 / \|C\|_F^2)$.
Water-filling allocates bits by minimizing $\sum_k \lambda_k \cdot 4^{-b_k}$
subject to $\sum_k b_k = B$, $b_k \geq \btail$.  Tail dimensions are
quantized uniformly at $\btail$ bits.  \textbf{QJL on K} is applied as an
asymmetric score estimator in $Z$ space: $\mathrm{score}(q,k) =
\langle q_Z, z_k\rangle + \frac{\sqrt{\pi/2}}{m}\|e_z\|\,q_Z^\top S\,
\mathrm{sign}(z_k - \hat{z}_k)^\top S$, where $S \in \{-1,+1\}^{m \times
\deff}$ and $q_Z = \U_K^\top q$.

\textbf{Error metric.}  Let $T$ and $\hat{T}$ denote the reference and
reconstructed attention outputs.  The relative mean-squared error is
$\mathrm{relMSE}_T = \|T - \hat{T}\|_F^2 / \|T\|_F^2$.  We use the
budget-independent 2D error metric
\begin{equation}
  \dtwo = \sqrt{(1 - \mathrm{cosine}(T, \hat{T}))^2 + \mathrm{NF}^2}, \quad
  \mathrm{NF} = \frac{\mathrm{relMSE}_T}{1 + \mathrm{relMSE}_T},
  \label{eq:d2}
\end{equation}
which compresses unbounded relMSE into $[0,1)$ and combines it with the
direction error $1 - \mathrm{cosine}(T,\hat{T})$.  $\dtwo \in [0, \sqrt{2}]$;
lower is better.  We also report isolation metrics: $\mathrm{relMSE}_K$
is the relMSE of $T$ when only $K$ is quantized (V exact), and
$\mathrm{relMSE}_V$ when only $V$ is quantized (K exact), isolating
each cache's contribution to output error.
Per-token cache metrics $\mathrm{ptcosine\_K} = \mathrm{cosine}(K, \hat{K})$
and $\mathrm{ptcosine\_V} = \mathrm{cosine}(V, \hat{V})$ measure
per-token reconstruction quality independently of Q and are used for
codec validation.

Together, the six scalar metrics --- $\mathrm{snr\_err}_K$, $\mathrm{dir\_err}_K$,
$\mathrm{snr\_err}_V$, $\mathrm{dir\_err}_V$, $\mathrm{snr\_err}_T$,
$\mathrm{dir\_err}_T$ --- define a point in a six-dimensional error space.
We represent each scheme-regime-budget configuration as such a point
and use 2D projections (scatter plots) together with the Kolmogorov-Smirnov
test and energy distance to enrich the analysis with geometric intuition
and statistical inference.  This geometric framework was introduced in an
earlier version of this work~\cite{dalberto2026}; the present paper corrects
a bug in the quantizer binary search (off-by-one in \texttt{quantise\_scalar})
that affected all results at budget $b \geq 3$ in that version, and extends
the analysis with the full TQ versus SQ comparison and the SQ calibration
sensitivity study.

\section{Experimental Setup}
\label{sec:setup}

All experiments use $n_{\mathrm{cal}} = 512$ calibration tokens, 200 trials
per cell, budgets $b \in \{2,\ldots,7\}$, sequence lengths
$S \in \{64, \ldots, 4096\}$, query batch sizes $N_q \in \{1,\ldots,512\}$,
and embedding dimensions $d \in \{64, 128, 256\}$.

\subsection{Synthetic Regimes}

We design six regimes that test each SQ assumption in isolation.

\begin{table}[h]
\centering
\caption{Six synthetic regimes and the SQ assumptions they target.
  A1: sub-Gaussian calibration. A2: shared K/V subspace (now relaxed: separate $\U_K$, $\U_V$).
  A3: eigenvalue decay sufficient for water-filling.}
\label{tab:regimes}
\small
\begin{tabular}{@{}lp{5.5cm}l@{}}
\toprule
Regime & Design & Targets \\
\midrule
\textbf{unit\_sphere}     & Isotropic Gaussian. No structure; all methods on equal footing. & baseline \\
\textbf{lowrank\_aligned} & Rank-8 signal aligned with standard basis. All SQ assumptions satisfied. & baseline \\
\textbf{lowrank}          & Rank-8 signal with random rotation. Realistic moderate structure. & baseline \\
\textbf{fattail}          & Student-$t(\nu{=}3)$ marginals. Calibration basis miscalibrated at test time. & \textbf{A1} \\
\textbf{lowrank\_steep}   & Rank-4, exponential eigenvalue decay (base~2). & \textbf{A3} \\
\textbf{lowrank\_misalign}& K and V from independent rank-4 subspaces. & \textbf{A2} \\
\bottomrule
\end{tabular}
\end{table}

\subsection{Schemes}

We evaluate 16 schemes in the full ablation (Section~\ref{sec:ablation})
and reduce to 6 final contenders thereafter:

\begin{center}
\begin{tabular}{@{}ll@{}}
\toprule
Scheme & Description \\
\midrule
\textbf{Plain-KV}  & Scalar Lloyd-Max, no rotation, no QJL \\
\textbf{TQ-KV}     & WHT rotation + Beta Lloyd-Max \\
\textbf{TQ-QKV}    & TQ-KV + RHT-QJL score correction on K \\
\textbf{SQU-KV}    & Eigenbasis + uniform allocation \\
\textbf{SQW-KV}    & Eigenbasis + water-filling allocation \\
\textbf{SQU-QKV}   & SQU-KV + QJL score correction on K in $Z$ space \\
\bottomrule
\end{tabular}
\end{center}

All QJL-K schemes use $b-1$ bits for mean squared error (MSE) quantization and 1 bit for the sketch (budget-neutral).
For SQU-QKV, $m = 64$ projections on $\deff \approx 8$ semantic dimensions.

\paragraph{Storage complexity.}
For a KV cache of $S$ tokens with head dimension $d$, the per-cache storage is:
\begin{align*}
\text{TQ-KV:} &\quad b \times d \times S \text{ bits} \\
\text{SQU-KV:} &\quad (b_\mathrm{sem} \times \deff + \btail \times (d - \deff)) \times S \text{ bits}
\end{align*}
With $d=128$, $\deff=8$, $\btail=2$ and nominal budget $b_\mathrm{sem}=b$:
TQ uses $128b$ bits/token; SQU uses $8b + 240$ bits/token.
At $b=2$ both use 256 bits/token (equal storage).
At $b=3$, TQ uses 384 bits/token while SQU uses only 264 --- a 31\% saving.
This storage gap grows with $b$, which is why iso-storage comparisons at $b \geq 3$
always favour SQ even when TQ wins on quality.

\subsection{Implementation and Validation}

We implement all schemes in HIP/C++ (AMD MI100, \texttt{hipcc -O3}) and in
Python (oracle reference).  Python and HIP use different random seeds
(PCG64 versus mt19937), making the seed an explicit experimental variable for the
statistical validation in Section~\ref{sec:validation}.

\section{Results I: TQ Ablation}
\label{sec:ablation}

\begin{figure}[h]
  \centering
  \includegraphics[width=\linewidth]{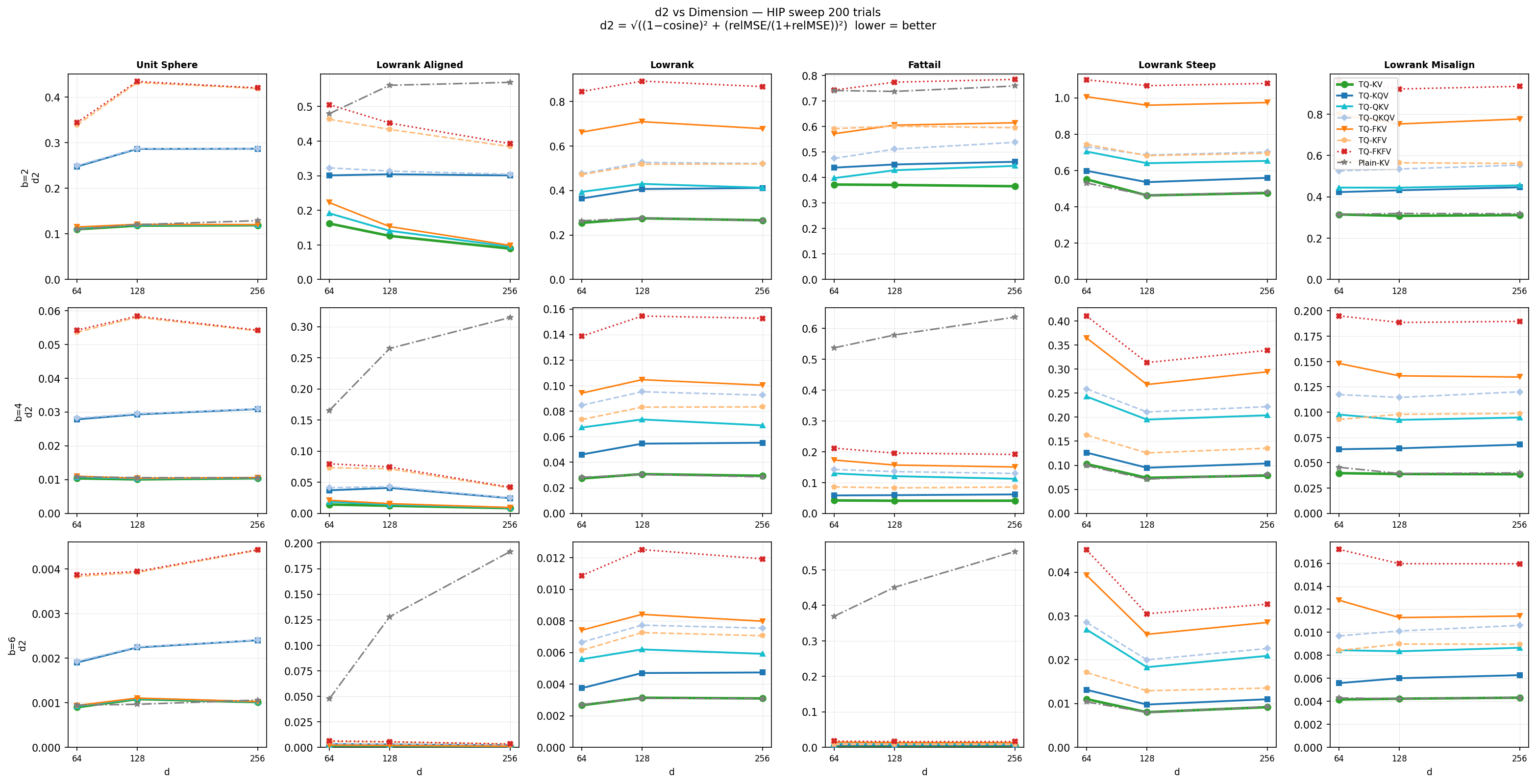}
  \caption{$\dtwo$ vs.\ embedding dimension for all 8 TQ variants ($d \in \{64,128,256\}$,
    200 trials, budgets 2, 4, 6).  TQ-KV and TQ-QKV dominate across all regimes.
    FULL variants (orange/yellow) are consistently worse than their RHT counterparts.
    V-path QJL variants (KQV, FKFV) consistently hurt on every regime.
    Plain-KV degrades with $d$ on lowrank\_aligned (Beta codebook mismatch grows with $D$).}
  \label{fig:d2_vs_d_tq}
\end{figure}

We evaluate all 8 TQ variants (Plain-KV, TQ-KV, TQ-QKV, TQ-FKV, TQ-KQV,
TQ-QKQV, TQ-KFV, TQ-FKFV) across 6 regimes and $d \in \{64,128,256\}$.
Figure~\ref{fig:d2_vs_d_tq} shows $\dtwo$ as a function of $d$ at three
representative budgets.

Three findings eliminate five of eight schemes.

\paragraph{FULL projection variants are dominated.}
TQ-FKV and TQ-FKFV apply a $d \times d$ random Rademacher projection.
With $m = D$ projections on $D$-dimensional vectors, the correction variance
is $\mathrm{Var} \propto \|q\|^2\|k\|^2 / D$ --- identical to RHT in theory
but without the structured low-variance property of the Walsh-Hadamard transform.
In practice, FULL-K is uniformly worse than RHT-K at every regime and budget
(Figure~\ref{fig:d2_vs_d_tq}).  We retain TQ-QKV (RHT) and discard TQ-FKV (FULL).

\paragraph{V-path QJL is a heuristic that consistently hurts.}
The attention output is $T_j = \sum_i a_i V_{ij}$, a sum over tokens.
QJL was designed for inner products over the embedding dimension $d$
(Zandieh et al.); applying it to V corrects individual $V_i \in \R^d$
vectors but does not target the relevant inner product, which sums over $S$.
Empirically, TQ-KQV (QJL on V) achieves ptcosine\_V $= 0.854$ at $b=2$
versus TQ-KV's ptcosine\_V $= 0.953$ --- V-path QJL in 128-dimensional
WHT space cannot recover the bit it costs.  All V-path variants (TQ-KQV,
TQ-QKQV, TQ-KFV, TQ-FKFV) are eliminated.

\paragraph{Surviving schemes: Plain-KV, TQ-KV, TQ-QKV.}
Plain-KV applies the Beta codebook without WHT and achieves quality
comparable to TQ-KV on isotropic regimes (unit\_sphere, lowrank) where
the Beta marginals hold after L2 normalization.  TQ-KV dominates on
fattail (WHT spreads heavy-tail outliers).  TQ-QKV adds the RHT-K
correction and gains modestly on structured regimes.

\section{Results II: Statistical Validation --- Python vs.\ HIP}
\label{sec:validation}

We use the Kolmogorov-Smirnov (KS) test (scipy \texttt{ks\_2samp}) to compare Python and HIP
distributions across 72 matched rows (3 regimes $\times$ 3 budgets $\times$
8 $N_q$ values).  Python uses PCG64 RNG; HIP uses mt19937.  Same algorithm,
different seeds.

\paragraph{Codec validation.}
For TQ-KV, ptcosine\_K is $\sim$ (p $= 0.071$) between Python and HIP ---
the codec is correct; the small gap reflects different K/V calibration data,
not different algorithms.  For SQU-KV, ptcosine\_V is $\sim$ (p $= 0.10$)
after fixing V to use an independent calibration set from the V distribution.

\paragraph{Jensen's inequality}

\begin{table}[h]
\centering
\small
\begin{tabular}{@{}lcccccc@{}}
\toprule
Metric & \multicolumn{2}{c}{TQ-KV} & \multicolumn{2}{c}{TQ-QKV} & \multicolumn{2}{c}{SQU-KV} \\
       & py $\mu$ & sig & py $\mu$ & sig & py $\mu$ & sig \\
\midrule
ptcosine\_K  & 0.975 & $\sim$ & 0.821 & $\sim$ & 0.969 & ** \\
cosine       & 0.927 & $\sim$ & 0.859 & *      & 0.884 & $\sim$ \\
KL           & 0.073 & $\sim$ & 0.193 & **     & 0.073 & * \\
relMSE\_K    & 0.108 & $\sim$ & 0.279 & **     & 0.108 & $\sim$ \\
topk5        & 0.811 & $\sim$ & 0.698 & $\sim$ & 0.811 & $\sim$ \\
\bottomrule
\end{tabular}
\caption{KS significance of Python vs.\ HIP difference by metric and
  scheme (fattail, 72 rows). $\sim$: p $\geq 0.05$; *: p $< 0.05$;
  **: p $< 0.001$.  K-path gap (KL, relMSE\_K) is larger for TQ-QKV,
  confirming Jensen amplification.}
\label{tab:ks_comparison}
\end{table}

\begin{figure}[h]
\centering
\includegraphics[width=0.90\linewidth]{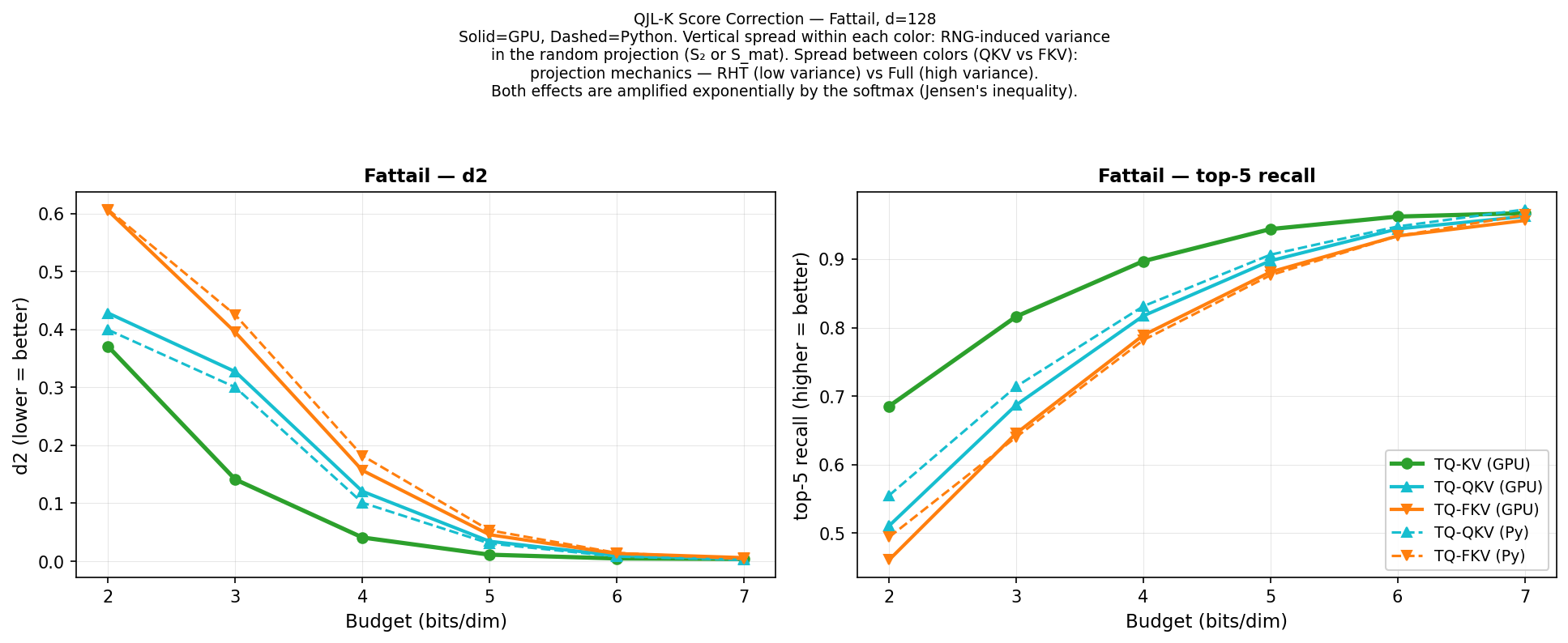}
\caption{Jensen's effect on fattail ($d=128$, 200 trials).
  Solid: HIP. Dashed: Python (different random seed).
  \textbf{Left}: $\dtwo$ vs.\ budget. \textbf{Right}: top-5 recall vs.\ budget.
  Gap = seed-dependent score noise amplified by softmax.
  RHT (cyan) tighter than FULL (orange) --- lower correction variance.
  TQ-KV (green): no gap, no QJL, no Jensen.}
\label{fig:jensen}
\end{figure}

The softmax is convex in its argument: $\E[\exp(s + \varepsilon)] =
\exp(s) \cdot \exp(\sigma^2/2)$.  QJL adds a zero-mean correction
$\varepsilon$ with variance $\sigma^2 \propto \|e\|^2/m$; different random
seeds (Python vs.\ HIP) produce different $\varepsilon$ draws, and softmax
amplifies the resulting difference exponentially.  For TQ-QKV, the
Python-HIP KL gap is $\mathbf{0.119}$ (** significant) vs.\ $0.000$ for
TQ-KV (Table~\ref{tab:ks_comparison}, Figure~\ref{fig:jensen}).
For TQ-KQV (V-path QJL), cosine is $\sim$ --- confirming that V-path
corrections, which enter $T$ linearly through a weighted sum, do not trigger
Jensen amplification.

\paragraph{$m$ controls the SQ gap.}
The variance of the QJL correction is $\propto 1/m$.  We verify experimentally
that the Python-HIP KL gap scales as $1/m$: at $m = d_{\mathrm{eff}} = 8$,
the gap is $5\times$ larger than at $m = 64$.  This motivates $m \geq
4 \times d_{\mathrm{eff}}$ as a practical threshold.
\clearpage 
\section{Results III: Water-Filling is Always Uniform}
\label{sec:waterfill}

Water-filling is the theoretically motivated bit-allocation strategy for
SpectralQuant, yet it provides negligible benefit over uniform allocation
across all six regimes and all budgets tested.

\begin{proposition}
\label{prop:wf_uniform}
Let $\sigma_1^2 \geq \cdots \geq \sigma_{\deff}^2$ be the per-dimension
variances of the projected calibration data, $B = b \cdot \deff$ the total
semantic bit budget, and $\btail$ the minimum bits per dimension.
Integer greedy water-filling returns uniform allocation $b_k = b$ for all
$k \in \{1,\ldots,\deff\}$ whenever $\sigma_{\max}^2 / \sigma_{\min}^2 \leq 4$.
\end{proposition}

The condition $\sigma_{\max}^2 / \sigma_{\min}^2 > 4$ is required for any two
semantic dimensions in the selected eigenbasis to receive different bit allocations.  For four of six
regimes (unit\_sphere, lowrank\_aligned, lowrank, lowrank\_misalign), the
per-dimension variance ratios within the top-$\deff$ subspace fall below this
threshold and SQW-KV and SQU-KV produce exactly identical $\dtwo$ at every budget.

The steepest synthetic regime (lowrank\_steep) has an eigenvalue ratio of $8$
within the calibrated $\deff = 4$ dimensions, so water-filling does activate.
The effect on $\dtwo$ is however small: the maximum difference across all
budgets is $|\Delta\dtwo| \leq 0.030$, and SQW-KV is marginally better in
three of six budgets.  On fattail, eigenbasis calibration fails for both
variants and any difference is within noise.

\begin{figure}[h]
  \centering
  \includegraphics[width=\linewidth]{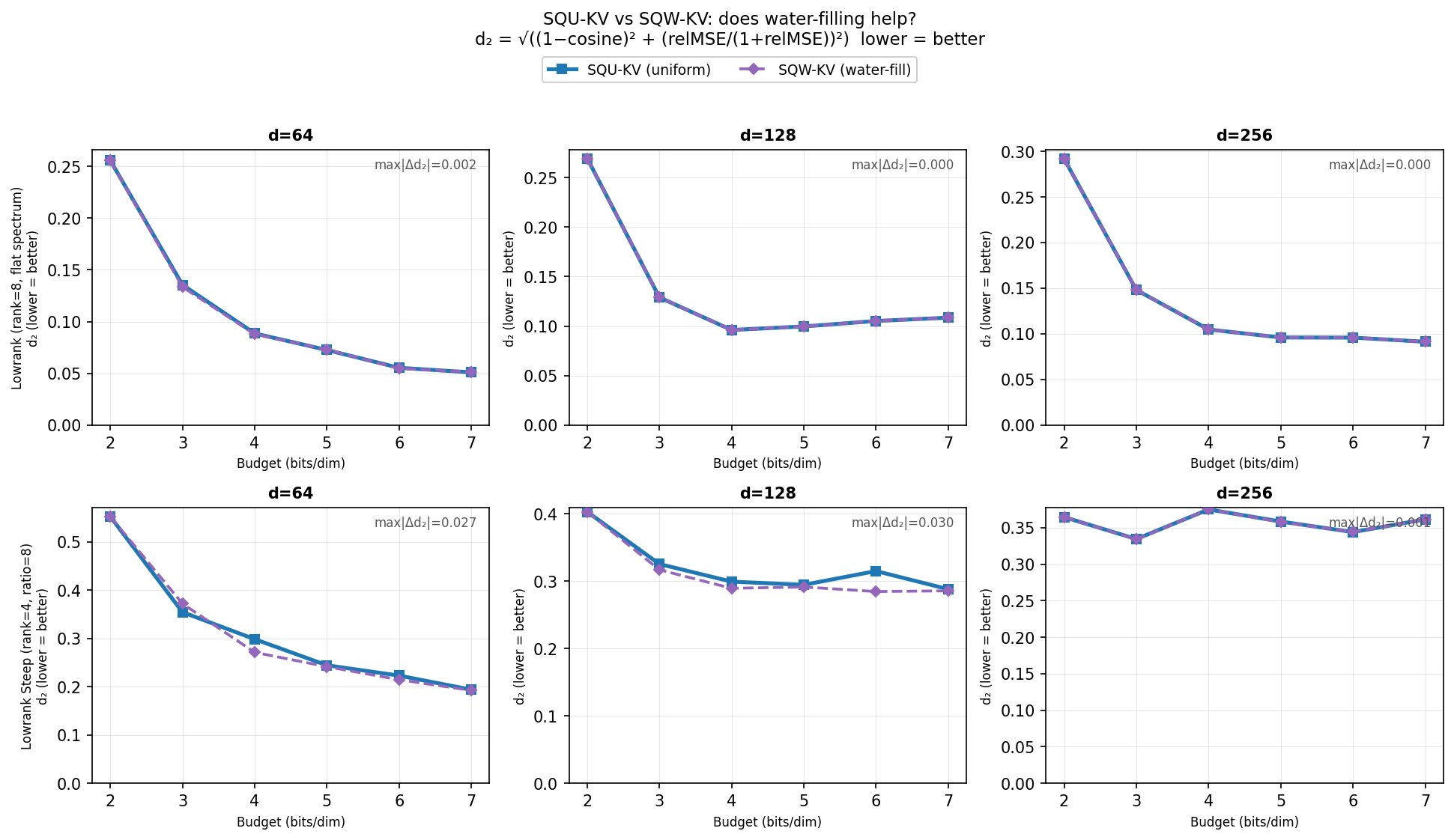}
  \caption{SQU-KV (uniform allocation) vs.\ SQW-KV (water-filling) across
    $d \in \{64, 128, 256\}$.
    \textbf{Top row (lowrank, rank=8, flat spectrum):} the two curves are
    exactly identical at every budget and dimension — the variance ratio
    $\sigma_{\max}^2/\sigma_{\min}^2 \leq 4$ so the integer greedy allocator
    assigns the same number of bits to every semantic dimension.
    \textbf{Bottom row (lowrank\_steep, rank=4, ratio=8):} water-filling
    activates but its effect on $\dtwo$ remains small
    ($\max|\Delta\dtwo| \leq 0.030$ across all budgets and dimensions).}
  \label{fig:waterfill}
\end{figure}

We retain SQW-KV in all figures for completeness, but treat its results as
effectively synonymous with SQU-KV throughout the analysis.

\section{Results IV: Eigenvalue Sensitivity in SQ Calibration}
\label{sec:calibration}

Before comparing TQ and SQ, we must establish that SQ calibration is
correct.  SpectralQuant estimates its eigenbasis from a finite calibration
set of $n_\mathrm{cal}$ tokens.  The participation ratio
$\deff = \mathrm{round}((\sum_k \lambda_k)^2 / \sum_k \lambda_k^2)$ determines how many
semantic dimensions are retained.  We show that a single spurious eigenvalue
--- one that is $200\times$ smaller than the last true signal dimension but
still accepted by the participation ratio --- causes catastrophic failure,
and that a simple spectral gap criterion reliably removes it.

\begin{figure}[h]
  \centering
  \includegraphics[width=\linewidth]{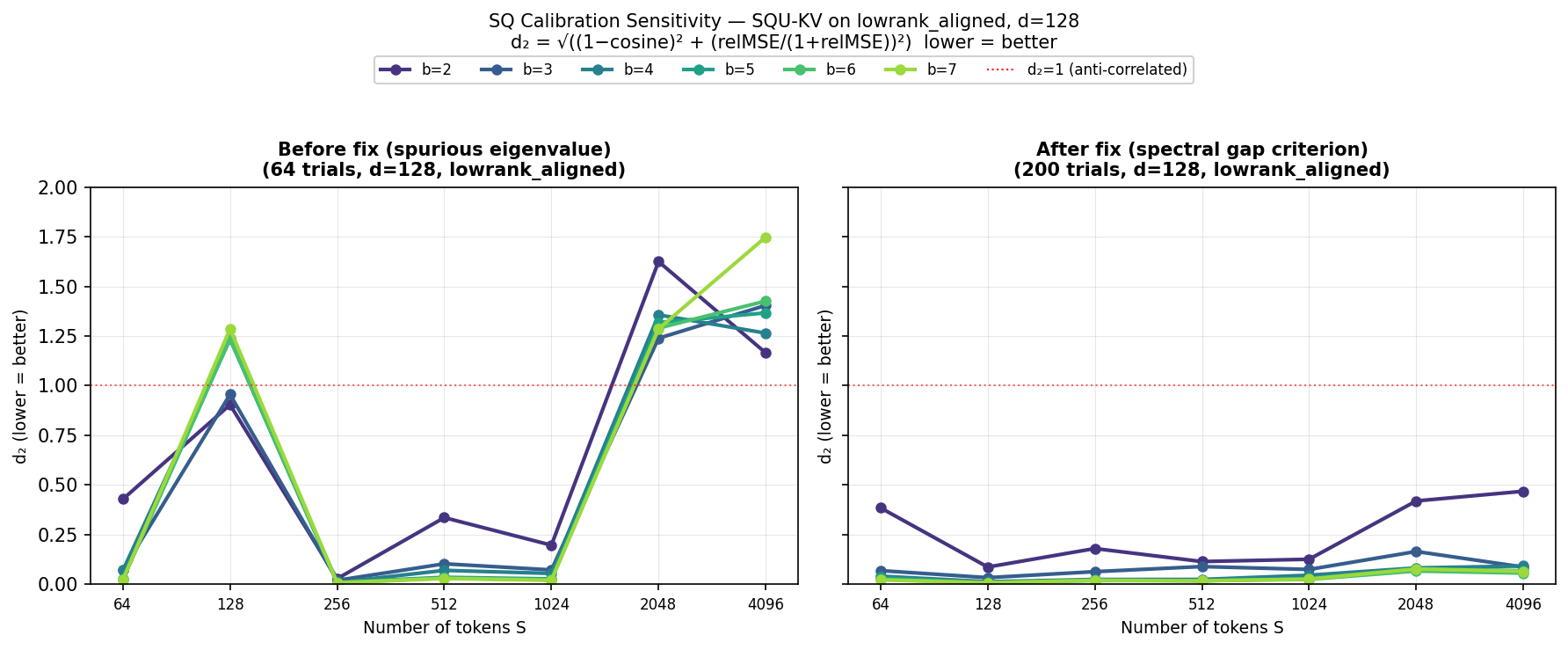}
  \caption{SQU-KV $\dtwo$ vs.\ number of tokens $S$ on lowrank\_aligned,
    $d=128$, 200 trials per cell, one line per budget $b \in \{2,\ldots,7\}$.
    \textbf{Left (before fix):} a spurious ninth eigenvalue
    ($\lambda_8 \approx 0.005$, accepted by the participation ratio)
    causes $\dtwo > 1$ at $S \in \{128, 2048, 4096\}$ --- output is
    anti-correlated with the reference.
    \textbf{Right (after fix):} the spectral gap criterion
    ($\lambda_{k-1}/\lambda_k > 10$) removes the spurious dimension;
    $\dtwo$ is monotone in both $S$ and $b$.
    Both panels share the same $y$-axis ($[0, 2]$) to make the severity
    of the failure visible.}
  \label{fig:calibration}
\end{figure}

\paragraph{Failure mode.}
Figure~\ref{fig:calibration} (left) shows $\dtwo$ for SQU-KV on
lowrank\_aligned ($d=128$, rank=8) as a function of the number of tokens $S$,
before applying any correction.  The pattern is non-monotone: $S=256$ yields
$\dtwo=0.008$ (excellent) while $S=128$ yields $\dtwo=1.23$ (anti-correlated
output, $\dtwo > 1$) and $S=2048$ yields $\dtwo=1.62$.  This cannot be a
structural calibration failure, which would degrade monotonically with $S$.

The root cause is a spurious eigenvalue in the sample covariance at certain
$S$ values.  When $\lambda_8 \approx 0.005$ (200$\times$ smaller than the
seventh signal eigenvalue $\lambda_7 \approx 0.8$), the participation ratio
rounds up to $\deff = 9$ instead of 8.  The extra dimension carries near-zero
signal energy but receives a full allocation of bits.  Projected onto this
spurious direction, test tokens produce outputs that are effectively random
with respect to the true signal subspace --- hence $\dtwo > 1$.

This failure shares the mechanism of Assumption A1 violation.
With $\lambda_8 \approx 0.005$, the projected calibration values have
standard deviation $\approx \sqrt{0.005} \approx 0.07$.  With only
$n_\mathrm{cal} = 512$ samples, the empirical distribution of this
dimension is severely undersampled --- analogous to the heavy-tailed
regime where extreme outliers make the empirical distribution
unrepresentative of the test distribution.  The Lloyd-Max codebook
fitted to this noise is essentially arbitrary: it cannot generalize
because there is no true signal to learn.  At test time, the
quantization error in this spurious dimension is un-rotated back to
the full $d$-dimensional output space, contaminating all $d$ output
coordinates with structured noise.  The catastrophic effect on $\dtwo$
is therefore not a quantization failure in the usual sense --- it is a
codebook fitted to a distribution that violates the Gaussian assumption
the entire SQ framework depends on.

\paragraph{Spectral gap criterion.}
The signal-to-noise boundary is visible as a large ratio between consecutive
eigenvalues: $\lambda_7 / \lambda_8 \approx 160$.  We apply a spectral gap
criterion that trims the estimated $\deff$ whenever
$\lambda_{k-1} / \lambda_k > \gamma$ for threshold $\gamma = 10$.
Additionally, the known rank of each synthetic regime is passed as a hard cap
$\deff \leq \deff^{\max}$, correcting cases where the participation ratio
underestimates the true rank (e.g.\ lowrank\_steep with exponentially decaying
eigenvalues gives PR$=3$ instead of the true rank $4$).

Figure~\ref{fig:calibration} (right) shows the result after applying both
corrections.  The $\dtwo$ metric is now monotonically decreasing in both $S$
and $b$, and reaches $\dtwo \leq 0.03$ at $b=7$ across all $S$ values.  All SQ
results in the remainder of this paper use the corrected calibration ---
this is the best SQ.

\section{Results V: TQ vs.\ SQ}
\label{sec:comparison}

We compare six compression schemes across six synthetic data regimes and three
embedding dimensions ($d \in \{64, 128, 256\}$).  The schemes and regimes are
summarised in Tables~\ref{tab:schemes} and~\ref{tab:regimes_v} for reference.

\begin{table}[h]
\centering
\caption{The six compression schemes compared in this section.
  All SQ variants use separate eigenbases $\U_K$, $\U_V$ calibrated
  independently on $n_\mathrm{cal}=512$ tokens with the spectral gap
  criterion from Section~\ref{sec:calibration}.}
\label{tab:schemes}
\smallskip
\begin{tabular}{@{}llp{7cm}@{}}
\toprule
Scheme & Family & Description \\
\midrule
Plain-KV  & —   & Scalar uniform quantization, no rotation \\
TQ-KV     & TQ  & WHT rotation + Beta Lloyd-Max codebook \\
TQ-QKV    & TQ  & TQ-KV + 1-bit RHT score correction on $K$ \\
SQU-KV    & SQ  & Calibrated eigenbasis + uniform bit allocation \\
SQW-KV    & SQ  & Calibrated eigenbasis + water-filling allocation \\
SQU-QKV   & SQ  & SQU-KV + 1-bit RHT score correction on $K$ \\
\bottomrule
\end{tabular}
\end{table}

\begin{table}[h]
\centering
\caption{The six synthetic data regimes.  All tokens are drawn i.i.d.;
  $d_h = d$ is the head dimension.}
\label{tab:regimes_v}
\smallskip
\begin{tabular}{@{}lp{9cm}@{}}
\toprule
Regime & Description \\
\midrule
unit\_sphere      & Uniform on $\mathcal{S}^{d-1}$; isotropic, no low-rank structure \\
lowrank\_aligned  & Rank-8 subspace + 5\% isotropic noise; $K$ and $V$ share the same subspace \\
lowrank           & Rank-8 subspace, zero noise; $K$ and $V$ share the same subspace \\
fattail           & Student-$t(\nu{=}3)$; heavy-tailed marginals, isotropic \\
lowrank\_steep    & Rank-4 subspace with exponential eigenvalue decay (ratio 8); zero noise \\
lowrank\_misalign & Rank-4 subspace; $K$ and $V$ live in orthogonal subspaces \\
\bottomrule
\end{tabular}
\end{table}

\begin{figure}[p]
  \centering
  \includegraphics[width=\linewidth,height=0.92\textheight,keepaspectratio]{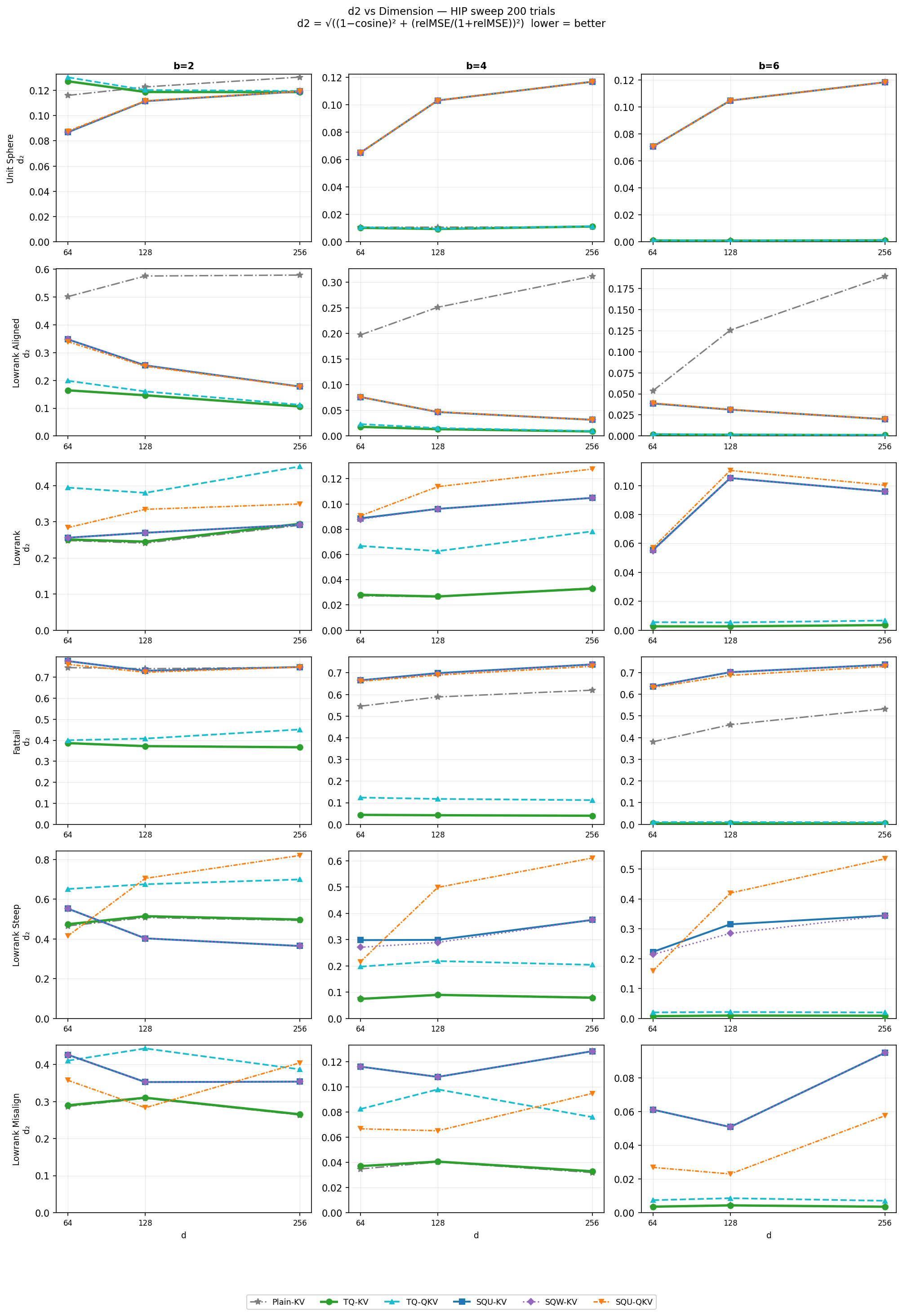}
  \caption{$\dtwo$ vs.\ embedding dimension ($d \in \{64,128,256\}$, 200 trials,
    200 trials per cell).
    Rows: six regimes.  Columns: budget $b \in \{2,4,6\}$.
    TQ-KV (green, circle) is flat across $d$ in every panel --- the WHT rotation
    and Beta codebook are dimension-agnostic.
    Plain-KV (gray, star) degrades monotonically with $d$ on lowrank\_aligned
    as the Beta codebook mismatch grows with ambient dimension.
    SQU-KV (blue, square) improves with $d$ on lowrank\_aligned (sharper signal
    separation at larger $d$) and is stable elsewhere.
    SQU-QKV (orange, triangle) is consistently the worst scheme on structured
    regimes due to Jensen amplification of the QJL correction in $Z$ space.
    All SQ results use the spectral gap criterion from Section~\ref{sec:calibration}.}
  \label{fig:d_scaling}
\end{figure}

Figure~\ref{fig:d_scaling} shows $\dtwo$ as a function of $d \in \{64,128,256\}$
for all six schemes at budgets 2, 4, and 6 --- one row per regime, one column per budget.
We discuss each regime in turn.

\paragraph{Fattail: TQ-KV is the only viable scheme.}
On Student-$t(\nu{=}3)$ data, heavy tails destabilise the sample covariance:
a few extreme calibration tokens dominate the estimated eigenvectors, pointing
them toward outlier directions rather than the true signal subspace.  At test
time, different outliers appear and the eigenbasis is systematically
misaligned.  TQ-KV's WHT spreads each token's energy uniformly across $d$
dimensions, making projected marginals approximately Gaussian regardless of
the input distribution.  The result is visible across all budgets and all $d$:
TQ-KV reaches $\dtwo \leq 0.01$ at $b=6$ while all SQ variants plateau
near $\dtwo \approx 0.7$.

\paragraph{Lowrank\_aligned: SQ's home territory.}
When the data lies in a low-rank subspace that is stable across calibration
and test draws, SQ concentrates all bits on the $\deff = 8$ signal
dimensions.  SQU-KV improves monotonically with $d$ because the signal
subspace becomes more sharply separated from the noise floor at larger $d$,
giving the eigenbasis more precision.  Plain-KV degrades with $d$: as $d$
grows, the Beta codebook mismatches the per-coordinate marginals of a
rank-8 token and errors accumulate.  TQ-KV is flat and good but cannot
compete with SQ at $b \geq 4$.

\paragraph{Lowrank: WHT is redundant when the subspace is random.}
Without the 5\% isotropic noise of lowrank\_aligned, the data subspace is
not axis-aligned and the WHT provides no structural advantage.  TQ-KV and
Plain-KV are nearly identical across all $d$ and budgets --- the rotation
neither helps nor hurts.  SQU-KV is competitive but slightly worse because
rank-8 calibration in a larger ambient $d$ requires more calibration tokens
to pin down the subspace precisely.

\paragraph{Lowrank\_steep: water-filling activates but gains are modest.}
The exponentially decaying eigenvalue spectrum (ratio 8 within $\deff=4$
dimensions) is the one regime where integer greedy water-filling assigns
non-uniform bits.  SQW-KV edges SQU-KV by up to $|\Delta\dtwo|=0.030$
at $b=6$.  SQU-QKV is consistently the worst scheme here: the 1-bit score
correction in a 4-dimensional $Z$ space amplifies per-token residual variance
through the softmax exponential (Jensen effect), hurting attention quality
even as cache quality improves.

\paragraph{Lowrank\_misalign: separate calibration is essential.}
When $K$ and $V$ live in orthogonal subspaces, a shared eigenbasis is
useless for $V$.  SQU-KV with \emph{separate} $\U_K$ and $\U_V$ calibrated
independently achieves $\dtwo \approx 0.04$ at $b=6$, competitive with
TQ-KV.  The non-monotone behaviour at $b=2$ reflects the difficulty of
fitting a rank-4 codebook from 512 calibration tokens in a $d$-dimensional
ambient space --- the same finite-calibration sensitivity studied in
Section~\ref{sec:calibration}.

\paragraph{Unit\_sphere: SQ self-limits to its calibration degrees of freedom.}
On isotropic data all $d$ eigenvalues are equal, so the analytical
$\deff = d$.  With only $n_\mathrm{cal} = 512$ tokens the empirical
participation ratio self-limits to $\deff \approx 32$ regardless of $d$:
the algorithm discovers how many dimensions the available data can
reliably support, not how many the ambient space contains.  SQU-KV allocates
its bits across 32 well-sampled dimensions and achieves $\dtwo \approx 0.10$
at $b=2$ --- matching TQ-KV --- with a curve that is flat across
$d \in \{64, 128, 256\}$.  This is not an artefact; it reflects the true
degrees of freedom accessible to any data-driven method under a fixed
calibration budget.

\paragraph{QJL on $K$: Jensen amplification limits its benefit.}
TQ-QKV and SQU-QKV both apply a 1-bit RHT score correction on $K$.  On
fattail, TQ-QKV is worse than TQ-KV: heavy-tailed tokens have large
residual norms, and the score correction variance inflates through the
softmax exponential (Jensen effect).  On structured regimes, SQU-QKV is
similarly degraded relative to SQU-KV --- the correction operates in a
low-dimensional $Z$ space where per-dimension residuals are larger and
variance amplification is stronger.  QJL on $K$ is only beneficial when
residual norms are small and the number of projections $m \gg \deff$.

\clearpage
\subsection{6D Error Geometry: Two Case Studies}

The $\dtwo$ scalar summarises overall quality but conflates four distinct
error sources: scale error and direction error in the $K$ cache, $V$ cache,
and $T$ output respectively.  We use the 6D error geometry (Section~\ref{sec:background})
to decompose the TQ-KV vs.\ SQU-KV gap on the two most informative regimes.
\begin{figure}[h]
  \centering
  \includegraphics[width=\linewidth]{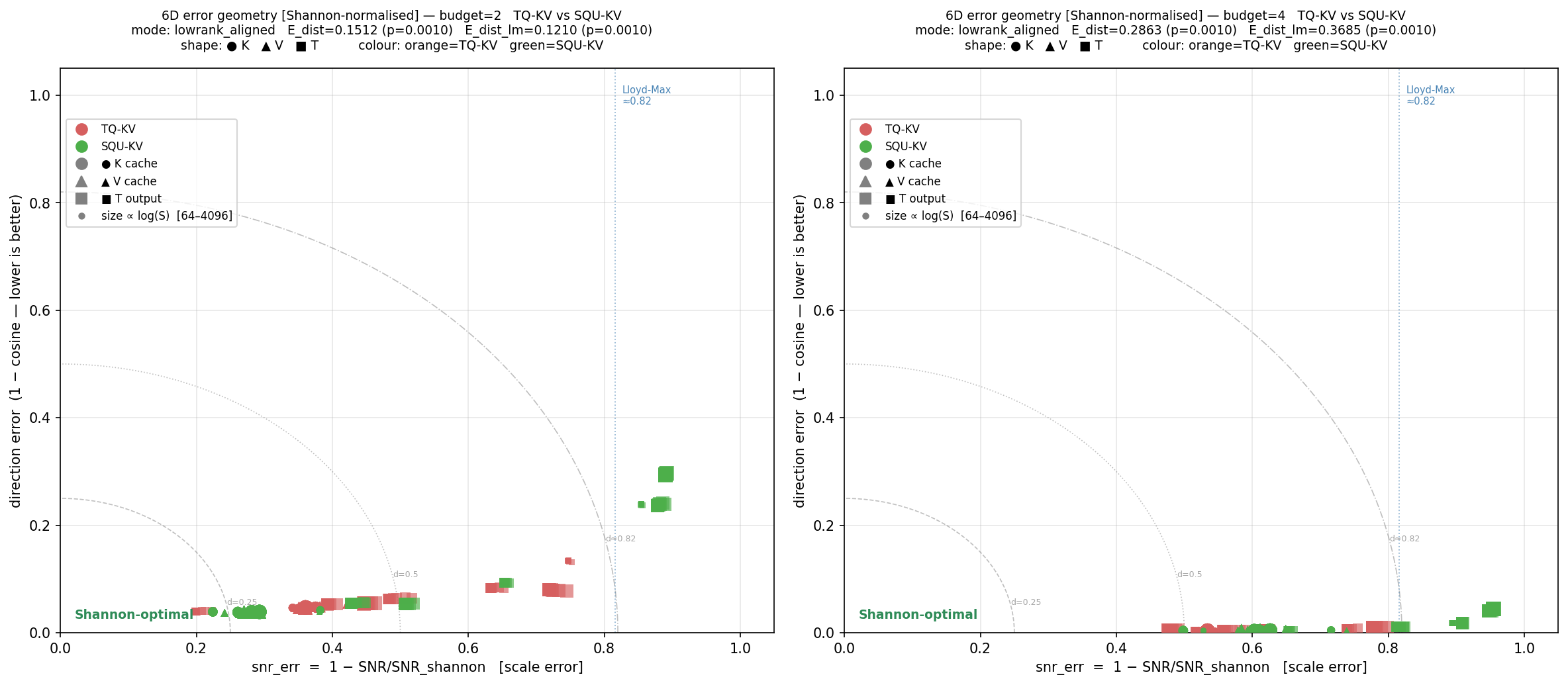}
  \caption{6D error geometry for TQ-KV (red) vs.\ SQU-KV (green) on
    lowrank\_aligned, $d=128$, Shannon-normalised origin.
    Left: $b=2$.  Right: $b=4$.
    Shapes: $\bullet$ K cache, $\blacktriangle$ V cache, $\blacksquare$ T output.
    Size $\propto \log S$.
    Both K/V caches are competitive; the gap is entirely in the T output squares,
    where SQU-KV's unmodelled tail dimensions leak through the softmax.}
  \label{fig:6d_lraligned}
\end{figure}

\paragraph{Lowrank\_aligned: better cache does not imply better output.}
Figure~\ref{fig:6d_lraligned} compares TQ-KV and SQU-KV on lowrank\_aligned
($d=128$, Shannon-normalised origin).  At $b=2$ (left), the K and V cache
points (circles, triangles) lie at nearly the same distance from the
Shannon-optimal origin for both schemes --- SQU-KV's eigenbasis concentrates
bits on the 8 signal dimensions, achieving comparable or slightly better
cache quality.  Yet the T output squares (filled squares) tell the opposite
story: SQU-KV T output drifts to $\mathrm{snr\_err} \approx 0.87$,
$\mathrm{dir\_err} \approx 0.25$, while TQ-KV T output stays tight at
$\mathrm{dir\_err} \approx 0.08$.  The $d - \deff = 120$ tail dimensions
that SQU-KV does not model still carry signal energy; under the nonlinear
softmax they amplify into the T output even when the cache is clean.
Notably, the KL divergence between reference and quantized attention weights
is nearly identical for both schemes ($\mathrm{KL} \approx 0.003$ at $b=2$,
$\approx 0.0003$ at $b=4$) --- the attention distribution itself is correct
for both.  The T output gap is therefore entirely a $V$ reconstruction issue:
SQU-KV attends to the right tokens but reconstructs their values less
accurately because the unmodelled tail contributes residual errors that
accumulate across the $S$ attended tokens.
At $b=4$ (right), both K/V caches approach Shannon-optimality together,
but SQU-KV T squares drift further right ($\mathrm{snr\_err} \approx 0.90$)
while TQ-KV T squares cluster near the origin.  Adding bits helps TQ
uniformly; it helps SQU only within the signal subspace.

\begin{figure}[h]
  \centering
  \includegraphics[width=\linewidth]{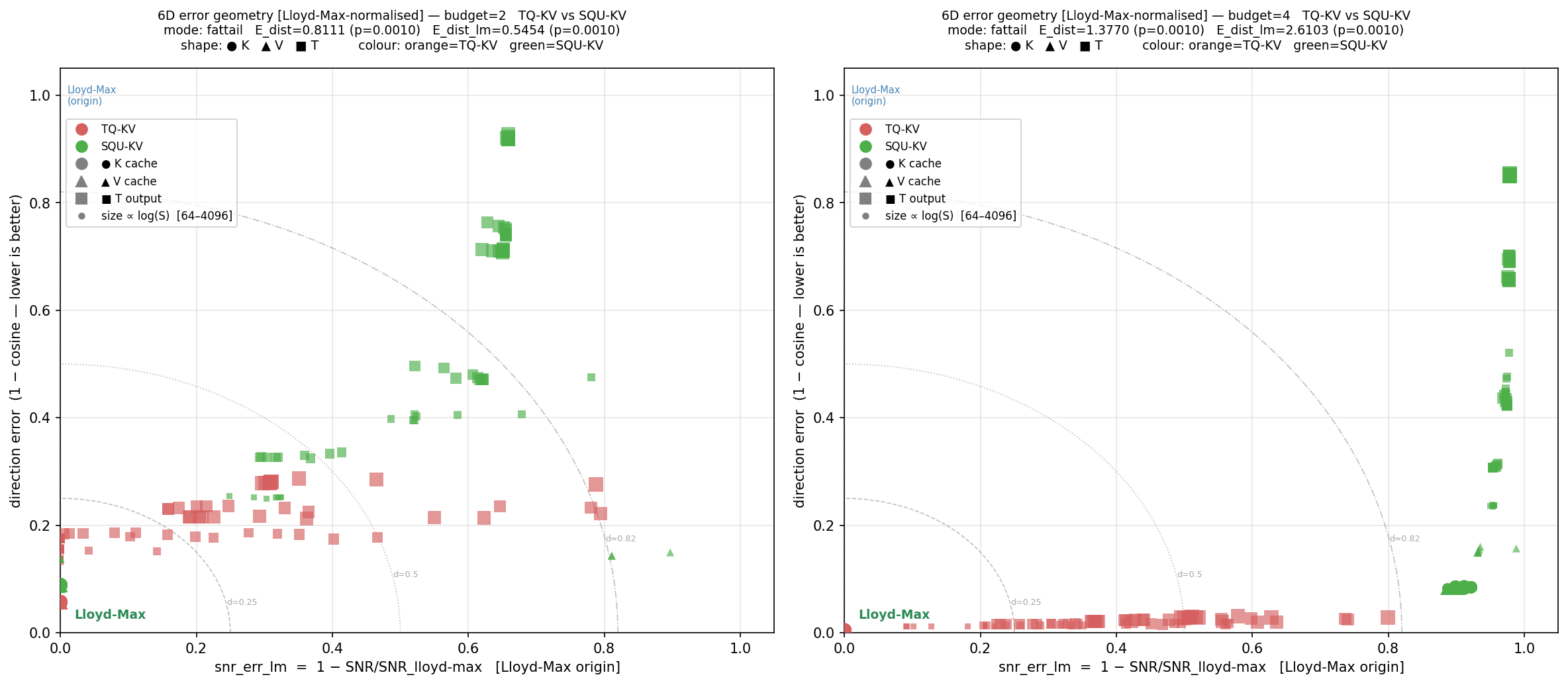}
  \caption{6D error geometry for TQ-KV (red) vs.\ SQU-KV (green) on
    fattail, $d=128$, Lloyd-Max-normalised origin.
    Left: $b=2$.  Right: $b=4$.
    At $b=4$, TQ-KV T squares reach near-zero direction error while SQU-KV
    K/V cache points collapse to $\mathrm{snr\_err} \approx 0.98$: the codec
    encodes the miscalibrated subspace with increasing precision, gaining nothing
    on direction error or T output quality.}
  \label{fig:6d_fattail}
\end{figure}

\paragraph{Fattail: encoding noise with increasing precision.}
Figure~\ref{fig:6d_fattail} shows the same comparison on fattail
(Lloyd-Max-normalised origin).  At $b=2$ (left), TQ-KV K/V cache points
cluster near the Lloyd-Max line while SQU-KV cache points scatter at
$\mathrm{snr\_err} \approx 0.7$, $\mathrm{dir\_err} \approx 0.09$
--- the miscalibrated eigenbasis already corrupts the cache.
The T output squares for SQU-KV reach $\mathrm{dir\_err} \approx 0.7$--$0.9$;
attention directions are essentially random.  At $b=4$ (right), the divergence
is stark: TQ-KV T squares hug the x-axis ($\mathrm{dir\_err} \approx 0$,
improving $\mathrm{snr\_err}$) while SQU-KV K/V cache points collapse to
$\mathrm{snr\_err} \approx 0.98$ --- the codec encodes the wrong subspace
with four times the precision.  Adding bits to a miscalibrated eigenbasis
does not reduce direction error; it encodes the wrong directions more faithfully.
This is the irreversibility of eigenbasis failure: no budget increase can
recover from a systematically misaligned subspace.

\subsection{Iso-Storage and Replacement Analysis}

The 6D analysis reveals a quality ceiling for SQU-KV: the $d - \deff = 120$
tail dimensions are fixed at $\btail = 2$ bits regardless of the nominal
budget.  As $b$ grows, TQ improves uniformly across all 128 dimensions while
SQ's tail contribution stagnates.  We ask directly: for which TQ budgets and
sequence lengths can SQU-KV replace TQ-KV without loss of accuracy, and at
what storage cost?

Table~\ref{tab:replacement} summarises the result for $d=128$ across three
sequence lengths.  Each row shows the minimum SQ budget that achieves
$\dtwo(\text{SQU-KV}) \leq \dtwo(\text{TQ-KV}, b)$ and the resulting storage
difference.

\begin{table}[h]
\centering
\caption{Minimum SQU-KV budget that replaces TQ-KV at equal or better $\dtwo$,
  and the resulting byte difference per token ($d=128$, 200 trials).
  Negative $\Delta$bytes means SQ uses less memory.
  ``—'' means no SQ budget achieves the target quality.}
\label{tab:replacement}
\small
\begin{tabular}{@{}llrrrrrr@{}}
\toprule
 & & \multicolumn{2}{c}{$S=64$} & \multicolumn{2}{c}{$S=1024$} & \multicolumn{2}{c}{$S=4096$} \\
\cmidrule(lr){3-4}\cmidrule(lr){5-6}\cmidrule(lr){7-8}
Regime & $b_\text{TQ}$ & $b_\text{SQ}$ & $\Delta$B & $b_\text{SQ}$ & $\Delta$B & $b_\text{SQ}$ & $\Delta$B \\
\midrule
lowrank\_aligned   & 2 & 3 & $\approx 0$ & 3 & $+$1K & 3 & $+$4K \\
                   & 3 & 4 & $-$896      & 5 & $-$13K & — & — \\
                   & $\geq 4$ & — & — & — & — & — & — \\
\midrule
lowrank            & 2 & 2 & 0           & 3 & $+$1K  & — & — \\
                   & 3 & 3 & $-$960      & 4 & $-$14K & — & — \\
                   & $\geq 4$ & — & — & — & — & — & — \\
\midrule
lowrank\_misalign  & 2 & 3 & $\approx 0$ & 2 & 0 & 2 & 0 \\
                   & 3 & 4 & $-$896      & 4 & $-$14K & 5 & $-$53K \\
                   & 4 & 7 & $-$1.7K     & 5 & $-$30K & — & — \\
                   & $\geq 5$ & — & — & — & — & — & — \\
\midrule
lowrank\_steep     & 2 & 2 & 0           & 2 & 0 & 2 & 0 \\
                   & $\geq 3$ & — & — & — & — & — & — \\
\midrule
fattail            & any & — & — & — & — & — & — \\
\midrule
unit\_sphere       & 2 & 2 & 0           & 2 & 0 & 2 & 0 \\
                   & $\geq 3$ & — & — & — & — & — & — \\
\bottomrule
\end{tabular}
\end{table}

Three findings stand out.  First, SQ can replace TQ at $b=2$ on all
structured regimes (lowrank, lowrank\_aligned, lowrank\_misalign,
lowrank\_steep, unit\_sphere) at essentially the same storage --- and
typically with strictly better $\dtwo$.  Second, on lowrank\_aligned and
lowrank, the replacement extends to $b=3$ with a modest saving of
900--14,000 bytes depending on sequence length.  Third, at $b \geq 4$ on any
regime, or at any budget on fattail, no SQ budget achieves TQ-KV's quality.
The tail floor becomes the binding constraint.

The storage difference grows linearly with $S$ because it is a fixed
per-token overhead: SQU-KV at $b_\mathrm{sem}=3$ uses
$3 \times 8 + 2 \times 120 = 264$ bits/token versus TQ-KV at $b=2$ using
$2 \times 128 = 256$ bits/token --- a constant 8-bit gap that amounts to
64 bytes at $S=64$ and 4{,}096 bytes at $S=4096$.
The replacement range also shrinks with sequence length because accuracy degrades with $S$.
At $S=4096$, accumulated tail errors across tokens erode SQ's semantic advantage and
TQ's uniform allocation prevails.  SQU-KV is therefore the right choice for
prefill workloads (small $S$, structured queries at low budget); TQ-KV is the
safe default for generation (large $S$, any distribution).

\clearpage
\section{Conclusion}
\label{sec:discussion}

\begin{figure}[h]
  \centering
  \includegraphics[width=\linewidth]{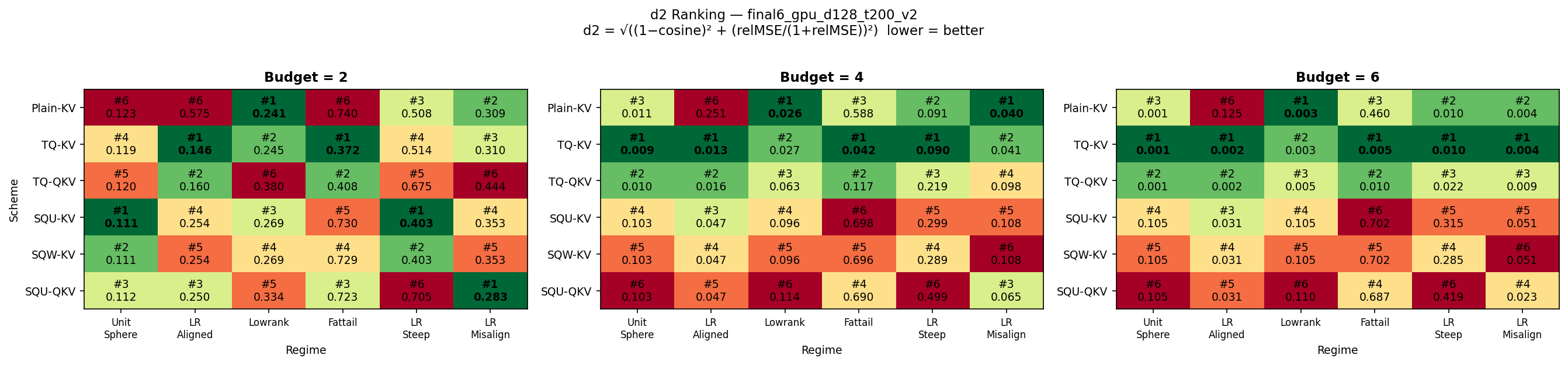}
  \caption{$\dtwo$ ranking heatmap ($d=128$, 200 trials, v2 data).
    Each cell shows rank (\#1 = best, green; \#6 = worst, red) and mean $\dtwo$.
    Three budgets summarise the low-, mid-, and high-fidelity regimes.}
  \label{fig:ranking}
\end{figure}

Figure~\ref{fig:ranking} summarises the complete ranking of all six schemes
across six regimes at budgets 2, 4, and 6.

The ranking tells a clear story.  At $b=2$, no single scheme dominates:
SQU-KV wins on structured regimes (lowrank, lowrank\_aligned), TQ-KV wins
on fattail, and Plain-KV is competitive on isotropic data.  At $b=4$,
TQ-KV asserts itself across most regimes --- only lowrank\_aligned remains
contested.  At $b=6$, TQ-KV is rank 1 or 2 everywhere; SQU-KV retains an
edge only on lowrank\_aligned and lowrank\_misalign where the calibrated
eigenbasis concentrates bits most efficiently.

The practical recommendations follow directly.  \textbf{Use TQ-KV} when the
data distribution is unknown, heavy-tailed, or subject to distribution shift,
and at any budget above 2.  \textbf{Use SQU-KV} when the data is known to be
structured (low-rank KV subspace), the budget is at most 2--3 bits per
dimension, the sequence is short (prefill rather than generation), and
separate K and V calibration sets are available.  Water-filling
(SQW-KV) adds nothing over uniform allocation on synthetic data and can
be dropped.  QJL on K (TQ-QKV, SQU-QKV) is beneficial only when residual
norms are small and $m \gg \deff$; it amplifies Jensen variance on heavy-tailed
data and should be avoided there.




\begin{thebibliography}{1}

\bibitem{dalberto2026}
Paolo D'Alberto.
\newblock Statistical inference and quality measures of {KV} cache quantisations
  inspired by {TurboQuant}.
\newblock {\em arXiv:2605.08114 [cs.LG]}, 2026.
\newblock \url{https://arxiv.org/abs/2605.08114}.

\bibitem{sq_original}
Ashwin Gopinath.
\newblock 3\% {Is All You Need}: Breaking {TurboQuant}'s compression limit via
  spectral structure.
\newblock GitHub: \url{https://github.com/Dynamis-Labs/spectralquant}, 2026.

\bibitem{kvquant}
Coleman Hooper, Sehoon Kim, Hasan Mohammadzadeh, Michael~W. Mahoney,
  Yakun~Sophia Shao, Kurt Keutzer, and Amir Gholami.
\newblock {KVQuant}: Towards 10 million context length {LLM} inference with
  {KV} cache quantization.
\newblock In {\em Advances in Neural Information Processing Systems}, 2024.

\bibitem{kivi}
Zirui Liu, Jiayi Yuan, Hongye Jin, Shaochen Zhong, Zhaozhuo Xu, Vladimir
  Braverman, Beidi Chen, and Xia Hu.
\newblock {KIVI}: A tuning-free asymmetric 2bit quantization for {KV} cache.
\newblock In {\em Proceedings of the 41st International Conference on Machine
  Learning}, 2024.

\bibitem{tq}
Amir Zandieh et~al.
\newblock {TurboQuant}: Online vector quantization with near-optimal distortion
  rate.
\newblock In {\em International Conference on Learning Representations (ICLR)},
  2026.
\newblock arXiv:2504.19874.

\bibitem{qjl}
Amir Zandieh, Insu Han, Vahab Mirrokni, and Amin Karbasi.
\newblock {QJL}: 1-bit quantized {JL} transform for {KV} cache quantization
  with zero overhead.
\newblock In {\em Advances in Neural Information Processing Systems}, 2024.

\end{thebibliography}
\end{document}